\documentclass{ifacconf}

\usepackage{natbib}
\usepackage{multicol}

\usepackage{amsmath}
\usepackage[T1]{fontenc}
\usepackage[latin9]{inputenc}
\usepackage{amsbsy}
\usepackage{amstext}
\usepackage{amssymb}
\usepackage{graphicx}
\usepackage{color}
\usepackage{pdflscape}
\usepackage{pifont}
\usepackage{booktabs,multirow}
\graphicspath{{./},{./fig/}}

\newif\ifjournalv
\journalvfalse

\ifjournalv\else

\fi

\usepackage[usenames,dvipsnames,svgnames,table]{xcolor}
\newcommand{\frmargin}[2]{{\color{Apricot}#1}\marginpar{\color{Apricot}\raggedright\footnotesize [FR]:
#2}}

\renewcommand{\frmargin}[2]{#1}

\newcommand{\BCM}{\textcolor{blue}{\ding{51}} }

\begin{document}
\begin{frontmatter}
\title{Review of Multi-Agent Algorithms for Collective Behavior: a Structural Taxonomy}


\author[First]{Federico Rossi} 
\author[Second]{Saptarshi Bandyopadhyay} 
\author[Second]{Michael Wolf}
\author[First]{Marco Pavone}

\address[First]{Department of Aeronautics and Astronautics, Stanford University (e-mail: \{frossi2, pavone\}@stanford.edu).}
\address[Second]{Jet  Propulsion  Laboratory,  California  Institute  of  Technology 
(e-mail: \{Saptarshi.Bandyopadhyay,michael.t.wolf\}@jpl.nasa.gov)}

   \thanks[footnoteinfo]{Part of this research was carried out at the Jet Propulsion Laboratory, California Institute of Technology, under a contract with the National Aeronautics and Space Administration. Federico Rossi and Marco Pavone were partially supported by the Office of Naval Research, Science of Autonomy Program, under Contract N00014-15-1-2673.
}



%

\maketitle

\begin{abstract}
In this paper, we  review multi-agent collective behavior algorithms in the literature and classify them according to their underlying mathematical structure. For each mathematical technique, we identify the multi-agent coordination tasks it can be applied to, and we analyze its scalability, bandwidth use, and demonstrated maturity. We highlight how versatile techniques such as artificial potential functions can be used for applications ranging from low-level position control to high-level coordination and task allocation, we discuss possible reasons for the slow adoption of complex distributed coordination algorithms in the field, and we highlight areas for further research and development. 
\end{abstract}


\end{frontmatter}

\ifjournalv\else \vspace{-1.8mm} \fi\section{Introduction}\ifjournalv\else \vspace{-1.8mm} \fi
\label{sec:intro}
%
Multi-agent robotic systems hold promise to enable new classes of missions in aerospace, terrestrial, and maritime
 applications, delivering higher resilience and adaptability at lower cost compared to existing monolythic systems. In particular, in the aerospace domain, multi-agent systems hold great promise for applications including multi-UAV patrolling, satellite formations for astronomy and Earth observation, and multi-robot planetary exploration. A number of algorithms have been proposed to control the collective behavior of such systems, ranging from low-level position control to high-level motion planning and  task allocation algorithms.

Many excellent surveys of algorithms for collective behavior exist in the literature; however, such papers generally focus either on single applications (e.g., formation control \citep{Ref:Oh15survey} or coverage \citep{Ref:Schwager09}) or on specific control techniques (e.g., consensus \citep{Ref:Garin10,Ref:Cao13}).
In contrast, in this paper, we survey the \emph{general} family of collective behavior algorithms for multi-agent systems and classify them according to their underlying mathematical \emph{structure}, without  restricting our focus to specific tasks or individual classes of algorithms.
In doing so, we aim to capture fundamental mathematical properties of algorithms (e.g. scalability with respect to the number of agents and bandwidth use) and to show how the same algorithm or family of algorithms can be applied to multiple tasks and missions.

In particular, the goal of this paper is threefold:
\begin{itemize}
\item to act as a guide to practitioners in the selection of control algorithms for a given task or application;
\item to highlight how mathematically similar algorithms can be used for a variety of tasks, ranging from low-level control to high-level coordination;
\item to explore the state-of-the-art in the field of control of multi-agent systems and identify areas for future research.
\end{itemize}
\emph{Tasks in multi-agent systems} 
can be broadly 
categorized
 into the following classes \citep{Ref:Brambilla13}: \\
(1) \textbf{Spatially-organizing behaviors}, where agents coordinate to achieve a given spatial configuration and have negligible interactions with the environment. These tasks can be further classified into: 
(a)~\textit{Aggregation:} converging to one location.
(b)~\textit{Pattern Formation:} achieving a desired formation. 
(c)~\textit{Coverage:} covering an area. \\
(2) \textbf{Collective explorations}, where agents interact with the environment but have minimal interaction among themselves. These tasks can be classified into:
(a)~\textit{Area Exploration:} exploring the environment for mapping or surveillance.
(b)~\textit{Goal Searching:} searching for targets. \\
(3) \textbf{Cooperative decision making}, where agents both coordinate among themselves and interact with the environment to accomplish complex tasks. These tasks can be further classified into:
(a)~\textit{Task Allocation:} distributing tasks among agents.
(b)~\textit{Collective Transport:} coordinating to transport large objects.
(c)~\textit{Motion Planning:} finding paths in cluttered environments.
(d)~\textit{Distributed Estimation:} estimating the state of one or multiple targets. \\ 
These simple tasks are the fundamental building blocks of many complex multi-agent applications. 

\rowcolors{2}{gray!25}{white}
\begin{table*}[htpb]
\centering
\begin{tabular}{l|lll|ll|llll||lll}
\toprule
{} & \rotatebox{90}{Aggregation} & \rotatebox{90}{Pattern Formation} & \rotatebox{90}{Coverage} & \rotatebox{90}{Area Exploration} & \rotatebox{90}{Goal Searching} & \rotatebox{90}{Task Allocation} & \rotatebox{90}{Collective Transport} & \rotatebox{90}{Motion Planning} & \rotatebox{90}{Distributed Estimation} & \rotatebox{90}{High Scalability} & \rotatebox{90}{Low Bandwidth Use} &    \rotatebox{90}{Maturity} \\
\midrule
\hline \textbf{Consensus}                                         &  \BCM &  \BCM &  \BCM &   &   &   &   &   &  \BCM &  \BCM &  \BCM &  \textcolor{ForestGreen}{H} \\
\hline \textbf{Artificial Potential Functions (APF)}              &  \BCM &  \BCM &  \BCM &  \BCM &   &  \BCM &  \BCM &  \BCM &   &  \BCM &  \BCM &  \textcolor{blue}{F} \\
\hline \textbf{Distributed Feedback Control}                      &  \BCM &  \BCM &   &   &   &   &   &   &  \BCM &  \BCM &  \BCM &  \textcolor{blue}{F} \\
\hline \textbf{Geometric Algorithms}                              &   &   &   &   &   &   &   &   &   &   &   &   \\
Voronoi-based Algorithms                                          &  \BCM &   &  \BCM &  \BCM &   &   &   &  \BCM &   &  \BCM &  \BCM &  \textcolor{ForestGreen}{H} \\
Circumcenter Algorithms                                           &  \BCM &  \BCM &   &   &   &   &   &   &   &  \BCM &  \BCM &  \textcolor{RedOrange}{S} \\
Bearing-only Algorithms                                           &  \BCM &  \BCM &   &   &   &   &   &   &   &  \BCM &  \BCM &  \textcolor{ForestGreen}{H} \\
Maze Searching Algorithms                                         &   &   &   &   &   &   &   &  \BCM &   &  \BCM &  \BCM &  \textcolor{RedOrange}{S} \\
Leader-Follower (LF) Algorithms                                   &   &  \BCM &   &   &   &   &   &   &   &  \BCM &  \BCM &  \textcolor{RedOrange}{S} \\
Velocity Obstacle (VO) based Algorithms                           &   &   &   &   &   &   &   &  \BCM &   &  \BCM &  \BCM &  \textcolor{blue}{F} \\
\hline \textbf{State Machines and Behavior Composition}           &   &   &   &   &   &   &   &   &   &   &   &   \\
Automata-based Algorithms                                         &   &   &   &   &   &  \BCM &   &   &  \BCM &  \BCM &  \BCM &  \textcolor{RedOrange}{S} \\
Behavior Composition                                              &   &   &   &   &   &  \BCM &  \BCM &   &   &   &   &  \textcolor{ForestGreen}{H} \\
Petri Networks                                                    &   &   &   &   &   &  \BCM &   &   &   &   &  - &  \textcolor{ForestGreen}{H} \\
Game Theory based Algorithms                                      &   &   &   &   &   &  \BCM &   &   &   &   &  - &  \textcolor{RedOrange}{S} \\
Resource Allocation Systems                                       &   &   &   &   &   &   &   &  \BCM &   &  \BCM &  - &  \textcolor{RedOrange}{S} \\
\hline \textbf{Bio-Inspired Algorithms}                           &   &   &   &   &   &   &   &   &   &   &   &   \\
Kilobot Self-Assembly Algorithm                                   &   &  \BCM &   &   &   &   &   &   &   &  \BCM &  \BCM &  \textcolor{ForestGreen}{H} \\
Optimotaxis Source-Searching Algorithm                            &   &   &   &   &  \BCM &   &   &   &   &  \BCM &  \BCM &  \textcolor{RedOrange}{S} \\
Beeclust Foraging Algorithm                                       &   &   &   &  \BCM &   &   &   &   &   &  \BCM &  \BCM &  \textcolor{RedOrange}{S} \\
Shepherding Algorithm                                             &  \BCM &   &   &   &   &   &   &   &   &  \BCM &  \BCM &  \textcolor{RedOrange}{S} \\
Termite-Inspired Collective Construction Algorithm                &   &   &   &   &   &  \BCM &  \BCM &   &   &  \BCM &  \BCM &  \textcolor{ForestGreen}{H} \\
Fish-inspired Goal Searching Algorithms                           &   &  \BCM &   &   &  \BCM &   &   &   &   &  \BCM &  \BCM &  \textcolor{ForestGreen}{H} \\
Gillespie Self-Assembly Algorithm                                 &   &  \BCM &   &   &   &   &   &   &   &  \BCM &  \BCM &  \textcolor{ForestGreen}{H} \\
Mergeable Modular Robots                                          &   &  \BCM &   &   &   &   &   &   &   &  \BCM &  \BCM &  \textcolor{ForestGreen}{H} \\
\hline \textbf{Density based Control}                             &   &   &   &   &   &   &   &   &   &   &   &   \\
Markov Chain-based Algorithms                                     &   &  \BCM &  \BCM &   &   &  \BCM &   &   &   &  \BCM &  \BCM &  \textcolor{ForestGreen}{H} \\
Smoothed Particle Hydrodynamics (SPH)                             &   &  \BCM &  \BCM &   &   &   &   &   &   &  \BCM &  \BCM &  \textcolor{ForestGreen}{H} \\
Optimal Transport based Algorithm                                 &   &  \BCM &   &   &   &   &   &  \BCM &   &  \BCM &  \BCM &  \textcolor{RedOrange}{S} \\
\hline \textbf{Distributed Optimization Algorithms}               &   &   &   &   &   &   &   &   &   &   &   &   \\
Distributed Linear Programming                                    &   &  \BCM &   &   &   &  \BCM &   &   &   &  \BCM &  \BCM &  \textcolor{RedOrange}{S} \\
Distributed Convex Optimization                                   &   &   \BCM &   &   &   & \BCM  &   &   &  \BCM &  \BCM &  \BCM &  \textcolor{RedOrange}{S} \\
Distributed Dynamic Programming                                   &   &   &   &   &   &  \BCM &   &  \BCM &   &   &   &  \textcolor{ForestGreen}{H} \\
Sequential Convex Programming                                     &   &   &   &   &   &   &   &  \BCM &   &  \BCM &  \BCM &  \textcolor{ForestGreen}{H} \\
Distributed Auction                                               &   &   &   &   &   &  \BCM &   &   &   &  \BCM &  \BCM &  \textcolor{ForestGreen}{H} \\
\hline \textbf{Local Optimization Algorithms for Global Behavior} &   &   &   &   &   &   &   &   &   &   &   &   \\
Decentralized Model Predictive Control (DMPC)                     &   &  \BCM &   &   &   &   &   &  \BCM &   &  \BCM &   &  \textcolor{ForestGreen}{H} \\
Formal Methods                                                    &   &   &   &   &   &   &   &  \BCM &   &  \BCM &  \BCM &  \textcolor{RedOrange}{S} \\
Sampling-based Motion-Planning Algorithms                         &   &   &   &   &   &   &   &  \BCM &   &   &   &  \textcolor{ForestGreen}{H} \\
\hline \textbf{Centralized Optimization Algorithms}               &   &   &   &   &   &   &   &   &   &   &   &   \\
MILPs and MINLPs                                                  &   &  \BCM &   &   &   &  \BCM &   &  \BCM & \BCM  &   &  - &  \textcolor{ForestGreen}{H} \\
Linear and Convex Optimization                                    &   &   &   &   &   &  \BCM &   &  \BCM &   \BCM&  \BCM &  - &  \textcolor{RedOrange}{S} \\
Markov Decision Processes (MDP)                                   &   &   &   &   &   &  \BCM &   &  \BCM &   &   &  - &  \textcolor{ForestGreen}{H} \\
Multi-Agent Traveling Salesman Problems                           &   &   &  \BCM &  \BCM &  \BCM &  \BCM &   &   &   &   &  - &  \textcolor{ForestGreen}{H} \\
Multi-Armed Bandits                                               &   &   &   &  \BCM &  \BCM & \BCM  &   &   &  \BCM &  \BCM &  - &  \textcolor{RedOrange}{S} \\
Direct Methods for Optimal Control                                &   &   &  \BCM &  \BCM &   &   &   &  \BCM &   &   &  - &  \textcolor{blue}{F} \\
Multiagent Reinforcement Learning                                 &   &   &   &  \BCM &   &  \BCM &   &   &   &   &  - &  \textcolor{ForestGreen}{H} \\
Frontier Techniques                                               &   &   &   &  \BCM & \BCM  &   &   &   &   &   &  - &  \textcolor{blue}{F} \\
Network Flow Algorithms                                           &   &   &   &   &   &  \BCM &   &  \BCM &   &  \BCM &  - &  \textcolor{RedOrange}{S} \\
Combinatorial Motion Planning                                     &   &   &   &   &   &   &   &  \BCM &   &  \BCM &  - &  \textcolor{RedOrange}{S} \\
\bottomrule
\end{tabular}
\caption{Categorization of collective behavior algorithms according to their mathematical structure and applicability of each algorithm to common multi-agent tasks. The scalability, bandwidth use, and level of demonstrated maturity of each algorithm (formally defined in Section \ref{sec:intro}) are also reported. }\label{tab:all_algos}
\end{table*}

\ifjournalv\else \vspace{-2.1mm} \fi
\subsubsection{Communication structure}  
In \textbf{centralized} algorithms, all agents share their information with a central node, which computes and issues a joint set of control actions. 
In \textbf{distributed} algorithms, agents can only explicitly share information with their neighbors. 
 Centralized algorithms can be implemented in a distributed fashion with a \textbf{shared-world} approach, discussed in Section \ref{sec:centralized-optimization-algo}.
 
\ifjournalv\else \vspace{-2.1mm} \fi
\subsubsection{Methodology}
We performed a thorough review of papers on multi-agent systems in major controls and robotics journals and conferences. 
It is not feasible to cite all existing works on control of multi-agent systems; accordingly, in this paper, we focus on identifying and classifying the key mathematical \emph{structures} and \emph{techniques}  that drive coordination algorithms, as opposed to individual contributions.  

We classify mathematical techniques according to their: 
(1)~\textbf{Scalability:} Highly scalable algorithms have been demonstrated on systems with more than 50 agents (in simulations or hardware). 
(2)~\textbf{Bandwidth use:} In low bandwidth algorithms, agents only communicate with their physical neighbors and do not exchange large messages. 
(3)~\textbf{Maturity:} The three classes of algorithms are: 
(i) only demonstrated in `simulation'~(\textcolor{RedOrange}{S})  
(ii) demonstrated in `hardware'~(\textcolor{ForestGreen}{H}) either in the lab or in technology demonstration missions
(iii) demonstrated in `field'~(\textcolor{Blue}{F}) deployments (excluding technology demonstrator missions).
\ifjournalv\else \vspace{-2.1mm} \fi
\subsubsection{Organization}
Our key contribution is Table \ref{tab:all_algos}, which reports the proposed taxonomy of mathematical techniques for collective behavior, highlights the tasks that each mathematical technique can achieve, and lists relevant performance metrics.
In Sections~\ref{sec:Consensus-algorithm}--\ref{sec:centralized-optimization-algo} we provide a synthetic description of the classification and relevant references. 
Finally, in Section~\ref{sec:conclusion} we draw conclusions and suggest directions for future research.

\ifjournalv\else \vspace{-1.8mm} \fi\section{A Structural Taxonomy of Multi-Agent Collective Behavior Algorithms}

\ifjournalv\else \vspace{-2.1mm} \fi\subsection{Consensus algorithms}\ifjournalv\else \vspace{-2.1mm} \fi\label{sec:Consensus-algorithm}
\textbf{Consensus} is among the oldest and most widely used distributed algorithms. Each agent shares and averages its state with its neighbors \ifjournalv\else\citep{Ref:Tsitsiklis86,Ref:Ren07}\fi. 
Applications include synchronization 
\ifjournalv\else \citep{Ref:Li2006global}\fi,
flocking \ifjournalv\else\citep{Ref:Tanner07,Ref:Saber06}\fi,
formation flying \ifjournalv\else\citep{Ref:Chung12}\fi, 
and distributed estimation \ifjournalv\else\citep{Ref:Rabbat04}\fi.
In \textit{gossip algorithms} \citep{Ref:Shah06}, each agent communicates with a single randomly-selected neighbor at each step.
In \textit{cyclic pursuit algorithms} \citep{Ref:Marshall04}, the consensus algorithm is executed on a directed ring communication topology. 


\ifjournalv\else \vspace{-2.1mm} \fi\subsection{Artificial Potential Functions (APF)}\ifjournalv\else \vspace{-2.1mm} \fi
\textbf{APF} algorithms synthesize agents' control inputs using the gradient of a suitably-defined potential function \citep{Ref:Khatib86}. 
These algorithms are very popular due to their simplicity, scalability, and ability to adapt to a number of tasks. Applications include pattern formation \ifjournalv\else \citep{Ref:Leonard07} \fi,
flocking \ifjournalv\else\citep{Ref:Zavlanos2007flocking}\fi,
 path planning \ifjournalv\else\citep{Ref:Koditschek90}\fi,
and task allocation \ifjournalv\else\citep{Ref:Weigel02}\fi.

\ifjournalv\else \vspace{-2.1mm} \fi\subsection{Distributed Feedback Control}\ifjournalv\else \vspace{-2.1mm} \fi
Each agent is endowed with a feedback controller that uses the agent's and its neighbors' states as the input
 \citep{Ref:Bamieh02,Ref:Feddema02}. In particular, tools for synthesis of \textbf{distributed LQG control} are available that can adapt to noisy communication links \citep{Ref:Sahai06}, and packet losses \ifjournalv\else \citep{Ref:Liu04}\fi, with applications to formation flying \citep{Ref:Ogren02} and distributed estimation.


\ifjournalv\else \vspace{-2.1mm} \fi\subsection{Geometric Algorithms}\ifjournalv\else \vspace{-2.1mm} \fi
In geometric algorithms, agents leverage their neighbors' location and speed information to perform spatially organizing tasks and path planning. 
\textbf{Voronoi algorithms} compute Voronoi partitions for coverage
\ifjournalv\else \citep{Ref:Bullo04}\fi,
path planning \ifjournalv\else \citep{Ref:Zhou17}\fi, 
and task allocation problems \citep{Ref:Pavone11}.
Other geometric algorithms include \textbf{circumcenter algorithms} for rendezvous  \ifjournalv\else\citep{Ref:Cortes06}\fi, 
\textbf{bearing-only algorithms} for formation control \citep{Ref:Fredslund02} and rendezvous \citep{Ref:Yu08rendezvous}, \textbf{maze searching algorithms} for  path planning \citep{Ref:Lumelsky97}, 
\textbf{leader-follower algorithms} for formation flying \ifjournalv\else\citep{Ref:Mesbahi99formation}\fi,
and \textbf{velocity obstacles} for collision avoidance
\ifjournalv\else\citep{Ref:vabdenBerg08}\fi.

\ifjournalv\else \vspace{-2.1mm} \fi\subsection{State Machines and Behavior Composition}\ifjournalv\else \vspace{-2.1mm} \fi
\textbf{Automata-based algorithms} leverage complex state machines and message-passing among agents to establish communication graphs and elect leaders for task allocation 
\ifjournalv\else \citep{Ref:Lynch97,Ref:Rossi14}\fi.
\textbf{Behavior composition} algorithms rely on composition of elementary behaviors for collective transport
\ifjournalv\else\citep{Ref:Rus1995moving}\fi.
\textbf{Petri networks} \ifjournalv\else\citep{Ref:King03} \fi
and \textbf{game theory} \citep{Ref:Arslan07} algorithms are used for centralized task allocation.
\textbf{Resource allocation systems} are used for multi-agent motion planning \citep{Ref:Reveliotis11}.

\ifjournalv\else \vspace{-2.1mm} \fi\subsection{Bio-Inspired Algorithms}\ifjournalv\else \vspace{-2.1mm} \fi
Bio-inspired algorithms mimic the behavior of swarms of animals such as insects and fish. 
We present a non-exhaustive list:
the \textbf{Kilobot algorithm} achieves complex two-dimensional shapes and was demonstrated on a thousand-agent testbed \citep{Ref:Nagpal14};
the \textbf{Optimotaxis source-searching algorithm} is inspired by the run and tumble behaviors of bacteria \citep{Ref:Hespanha08};
 the  \textbf{Beeclust foraging algorithm} is inspired by the behavior of honey bees \citep{Ref:Hereford11};
\textbf{Shepherding algorithms} enable control of large numbers of uncontrolled agents with few controlled agents \citep{Ref:Strobom14};
a \textbf{Termite-inspired algorithm} generates low-level rules for construction of complex structures \ifjournalv\else\citep{Ref:Nagpal14termite}\fi;
a \textbf{Fish-inspired goal-searching algorithm} switches between individual and collective behavior based on confidence level \citep{Ref:Wu12};
the \textbf{Gillespie self-assembly algorithm} leverages chemical kinetics;
\textbf{Mergeable modular robots} connect to form larger bodies or split into separate bodies, with self-healing properties \ifjournalv\else\citep{Ref:Dorigo17}\fi.

\ifjournalv\else \vspace{-2.1mm} \fi\subsection{Density based Control}\ifjournalv\else \vspace{-2.1mm} \fi
As opposed to the agent-based \emph{Lagrangian} framework, density-based algorithms  adopt an \emph{Eulerian} framework by treating agents as a continuum and controlling their density.  
\textbf{Markov chain} based algorithms partition the workspace into disjoint cells and control the transition probabilities between cells for pattern formation and goal searching applications \citep{Ref:Acikmese12,Ref:Bandyopadhyay17_TRO}.
\textbf{Smoothed particle hydrodynamics (SPH)} \citep{Ref:MKumar11} and \textbf{optimal transport} \citep{Ref:Bandyopadhyay14MSC}  based algorithms are also used for swarm formation control.

\ifjournalv\else \vspace{-2.1mm} \fi\subsection{Distributed Optimization Algorithms}\ifjournalv\else \vspace{-2.1mm} \fi
Distributed optimization algorithms allow agents to jointly solve optimization problems through information exchange and local computations. 
\textbf{Distributed linear programming} \ifjournalv\else\citep{Ref:Bullo12} \fi 
is used for pattern formation and task allocation; \textbf{distributed convex optimization} can encode richer convex constraints \citep{Ref:Boyd11}. \textbf{Distributed dynamic programming} \citep{Ref:Bertsekas1982distributed} is used for task allocation and motion planning.
\textbf{Sequential Convex Programming} can solve non-convex motion planning problems through local convexification and iteration \citep{Ref:Morgan15_SATO}.
The above algorithms can also be used in a \textbf{distributed model-predictive control} framework \citep{Ref:Scattolini09}. 
Market-based protocols like \textbf{distributed auction} \ifjournalv\else\citep{Ref:Gerkey02}\fi, mechanism design \ifjournalv\else\citep{Ref:Dias04}\fi, and coalition formation \citep{Ref:Shehory98} are widely used for task allocation.

\ifjournalv\else \vspace{-2.1mm} \fi\subsection{Local optimization algorithms for global behavior}\ifjournalv\else \vspace{-2.1mm} \fi
\label{sec:local-optimization-algo}
In local optimization algorithms, each agent solves an optimization problem; while the resulting behavior is not generally optimal for the entire system, favorable global properties such as collision avoidance can be guaranteed.
In \textbf{decentralized model predictive control (DMPC)}
each agent employs a local model-predictive control algorithms; inter-agent communication is used to coordinate the agents' plans \ifjournalv\else\citep{Ref:How07MPC}\fi. Distributed MPC has been used for flocking and motion planning \ifjournalv\else\citep{Ref:Dunbar02,Ref:How06}\fi.
\textbf{Formal methods} are used in concert with low-level control primitives for multi-agent motion planning with guaranteed collision avoidance \citep{Ref:KressGazit08}.
Decentralized multi-agent \textbf{sampling-based motion planning algorithms} have enjoyed significant practical success because of their ease of implementation, ability to handle higher-dimensional spaces, probabilistic completeness, and asymptotic optimality \ifjournalv\else\citep{Ref:Bandyopadhyay17_MA_SESCP,Ref:How12}\fi.

\ifjournalv\else \vspace{-2.1mm} \fi\subsection{Centralized optimization algorithms}\ifjournalv\else \vspace{-2.1mm} \fi \label{sec:centralized-optimization-algo}
\textbf{Mixed-integer linear programs (MILPs)} and mixed-integer convex programs (MICPs), can 
 solve simultaneous task allocation and path planning \citep{Ref:How03}, tracking \citep{Ref:Zhe13}, formation flying \citep{Ref:How02}, and defend-the-flag problems \citep{Ref:Earl02modeling}.
\textbf{Linear and convex optimization} problems can also be used to solve task allocation problems  \citep{Ref:Bertsekas98b,Ref:Kumar14} with  collision avoidance constraints \citep{Ref:Acikmese06convex}, and for distributed estimation and target tracking \citep{Ref:Aslam2003tracking}.
\textbf{Markov decision processes (MDPs)} and partially observable MDPs capture the stochastic nature of the environment and model the agents' \emph{coordination mechanism} \citep{Ref:Boutilier99b}
. POMDPs have been used for multi-agent path planning \citep{Ref:Ali15} and task allocation. 
Several approximation algorithms are available to solve the \textbf{m-vehicle traveling salesman problem (TSP)} and the team orienteering problem, building blocks for spatial task allocation, persistent monitoring, and information-gathering problems \citep{Ref:Rus14}.
\textbf{Multi-agent multi-armed bandit} problems \citep{Ref:Gittins79} capture the trade-off between exploration and exploitation: they have been employed for task allocation \ifjournalv\else\citep{Ref:Le08} \fi, goal searching, and tracking applications \citep{Ref:Leonard16}.
\textbf{Direct methods for trajectory optimization} \citep{Ref:VonStryk92} are used for area coverage, goal searching, and motion planning \citep{Ref:Leonard10b}.
\textbf{Multi-agent reinforcement learning (MARL)}
 has been used for exploration \citep{Ref:Chalkiadakis03} and task allocation \citep{Ref:Liu16}.
\textbf{Frontier techniques} \ifjournalv\else\citep{Ref:Burgard00} \fi are used for  urban search-and-rescue, reconnaissance \citep{Ref:Olson12} and sample collection \citep{Ref:Eich14}.
\ifjournalv
\else \textbf{Network flow}  \fi formulations have been proposed for Air Traffic Control \citep{Ref:Menon04} and for control of  autonomous vehicles offering on-demand transportation \citep{Ref:Pavone11b, Ref:Rossi17a}. 
Several \textbf{cooperative combinatorial motion planning} algorithms have been proposed for multi-agent systems: we refer the reader to \citep{Ref:Sturtevant15} for a thorough review. 
Centralized optimization algorithms can be implemented in a distributed fashion with a \textbf{shared-world} approach, where agents exchange their state and observations so that every robot has full knowledge of the entire system's state. However, shared-world algorithms have very onerous communication requirements (due to large messages and all-to-all communication) and high computation complexity, since each agent must solve the full centralized optimization problem.

\ifjournalv\else \vspace{-1.8mm} \fi\section{Conclusion}\ifjournalv\else \vspace{-1.8mm} \fi \label{sec:conclusion}

The proposed taxonomy and the properties shown in Table \ref{tab:all_algos} highlight some surprising characteristics of collective behavior algorithms.
The majority of existing mathematical techniques is tailored to either low-level spatially organizing tasks (e.g., bio-inspired algorithms and density-based control) or high-level coordination applications (e.g., state machines and optimization-based algorithms).
Only a small number of mathematical techniques (in particular, Artificial Potential Functions) can be adapted to a wide variety of tasks that include both low-level and high-level application.
This prompts further research into non-APF algorithms for multi-agent systems that share APF's key properties of simplicity, scalability, and high expressivity. 

Very few algorithms are mature and field-tested. 
Such algorithms exchange very simple information (e.g. the agents' locations) or rely on centralized implementations: this may be justified by the difficulty of characterizing and certifying the behavior of an entire multi-agent system when distributed algorithms are used. To overcome this, (i) research in formal methods and adoption of tools from the distributed algorithms literature to provide stronger guarantees for distributed systems and (ii) creation of standardized software and hardware test-beds to characterize the end-to-end behavior of such systems are needed.


Several avenues for future research are of interest. In particular, we hope to evaluate the performance of collective behavior algorithms according to additional metrics including 1) bandwidth use in broadcast and in point-to-point networks, 2) computational complexity, 3) availability of formal guarantees, 4) resilience to disruptions in communication network and to \emph{adversarial} failures, and 5) availability of a reference implementation.
We also wish to explore other possible taxonomies for coordination algorithms based, e.g., on the content of messages exchanged by the agent (which vary from simple ``beacon'' messages reporting the agent's location to complex messages carrying intentions and bids), and the communication topology induced by the algorithm (single-hop vs. multi-hop)
Finally, we plan to further explore high-level multi-agent tasks, including adversarial ``swarm vs. swarm'' problems, and to assess the applicability and performance of collective behavior algorithms with respect to such tasks.



{\small
\ifjournalv\else \vspace{-2mm} \fi
\bibliography{SurveyBib}

\begin{thebibliography}{89}
\providecommand{\natexlab}[1]{#1}
\providecommand{\url}[1]{\texttt{#1}}
\providecommand{\urlprefix}{URL }
\expandafter\ifx\csname urlstyle\endcsname\relax
  \providecommand{\doi}[1]{doi:\discretionary{}{}{}#1}\else
  \providecommand{\doi}{doi:\discretionary{}{}{}\begingroup
  \urlstyle{rm}\Url}\fi

\bibitem[{A\c{c}ikme\c{s}e and Bayard(2012)}]{Ref:Acikmese12}
A\c{c}ikme\c{s}e, B. and Bayard, D.S. (2012).
\newblock A {M}arkov chain approach to probabilistic swarm guidance.
\newblock In \emph{{IEEE American Control Conf.}}

\bibitem[{A{\c{c}}{\i}kmese et~al.(2006)A{\c{c}}{\i}kmese, Scharf, Murray, and
  Hadaegh}]{Ref:Acikmese06convex}
A{\c{c}}{\i}kmese, B., Scharf, D.P., Murray, E.A., and Hadaegh, F.Y. (2006).
\newblock A convex guidance algorithm for formation reconfiguration.
\newblock In \emph{{AIAA Guidance, Navigation, and Control Conference}}.

\bibitem[{Arslan et~al.(2007)Arslan, Marden, and Shamma}]{Ref:Arslan07}
Arslan, G., Marden, J.R., and Shamma, J.S. (2007).
\newblock Autonomous vehicle-target assignment: A game theoretical formulation.
\newblock \emph{{ASME} J. Dyn. Syst. Meas. Control}, 129(5), 584--596.

\bibitem[{Aslam et~al.(2003)Aslam, Butler, Constantin, Crespi, Cybenko, and
  Rus}]{Ref:Aslam2003tracking}
Aslam, J., Butler, Z., Constantin, F., Crespi, V., Cybenko, G., and Rus, D.
  (2003).
\newblock Tracking a moving object with a binary sensor network.
\newblock In \emph{Proceedings of the 1st international conference on Embedded
  networked sensor systems}.

\bibitem[{Bamieh et~al.(2002)Bamieh, Paganini, and Dahleh}]{Ref:Bamieh02}
Bamieh, B., Paganini, F., and Dahleh, M.A. (2002).
\newblock Distributed control of spatially invariant systems.
\newblock \emph{{IEEE} Trans. Autom. Control}, 47(7), 1091--1107.

\bibitem[{Bandyopadhyay et~al.(2017{\natexlab{a}})Bandyopadhyay, Baldini,
  Foust, Rahmani, de~la Croix, Chung, and
  Hadaegh}]{Ref:Bandyopadhyay17_MA_SESCP}
Bandyopadhyay, S., Baldini, F., Foust, R., Rahmani, A., de~la Croix, J.P.,
  Chung, S.J., and Hadaegh, F.Y. (2017{\natexlab{a}}).
\newblock Distributed fast motion planning for spacecraft swarms in cluttered
  environments using spherical expansions and sequence of convex optimization
  problems.
\newblock In \emph{9th International Workshop on Satellite Constellations and
  Formation Flying}.

\bibitem[{Bandyopadhyay et~al.(2014)Bandyopadhyay, Chung, and
  Hadaegh}]{Ref:Bandyopadhyay14MSC}
Bandyopadhyay, S., Chung, S.J., and Hadaegh, F.Y. (2014).
\newblock Probabilistic swarm guidance using optimal transport.
\newblock In \emph{Proc. {IEEE} Conf. Control Applicat.}

\bibitem[{Bandyopadhyay et~al.(2017{\natexlab{b}})Bandyopadhyay, Chung, and
  Hadaegh}]{Ref:Bandyopadhyay17_TRO}
Bandyopadhyay, S., Chung, S.J., and Hadaegh, F.Y. (2017{\natexlab{b}}).
\newblock Probabilistic and distributed control of a large-scale swarm of
  autonomous agents.
\newblock \emph{{IEEE} Trans. Robotics}, 33(3).

\bibitem[{Bellingham et~al.(2003)Bellingham, Tillerson, Richards, and
  How}]{Ref:How03}
Bellingham, J., Tillerson, M., Richards, A., and How, J.P. (2003).
\newblock Multi-task allocation and path planning for cooperating uavs.
\newblock In \emph{Cooperative Control: Models, Applications and Algorithms}.
  Springer.

\bibitem[{Bertsekas(1982)}]{Ref:Bertsekas1982distributed}
Bertsekas, D. (1982).
\newblock Distributed dynamic programming.
\newblock \emph{{IEEE} Trans. Autom. Control}, 27(3), 610--616.

\bibitem[{Bertsekas(1998)}]{Ref:Bertsekas98b}
Bertsekas, D.P. (1998).
\newblock \emph{Network optimization: continuous and discrete models}.
\newblock Athena Scientific.

\bibitem[{Boutilier(1999)}]{Ref:Boutilier99b}
Boutilier, C. (1999).
\newblock Sequential optimality and coordination in multiagent systems.
\newblock In \emph{IJCAI}, volume~99.

\bibitem[{Boyd et~al.(2006)Boyd, Ghosh, Prabhakar, and Shah}]{Ref:Shah06}
Boyd, S., Ghosh, A., Prabhakar, B., and Shah, D. (2006).
\newblock Randomized gossip algorithms.
\newblock \emph{{IEEE} Trans. Inf. Theory}, 52, 2508 -- 2530.

\bibitem[{Boyd et~al.(2011)Boyd, Parikh, Chu, Peleato, and
  Eckstein}]{Ref:Boyd11}
Boyd, S., Parikh, N., Chu, E., Peleato, B., and Eckstein, J. (2011).
\newblock Distributed optimization and statistical learning via the alternating
  direction method of multipliers.
\newblock \emph{Foundations and Trends in Machine Learning}, 3(1), 1--122.

\bibitem[{Brambilla et~al.(2013)Brambilla, Ferrante, Birattari, and
  Dorigo}]{Ref:Brambilla13}
Brambilla, M., Ferrante, E., Birattari, M., and Dorigo, M. (2013).
\newblock Swarm robotics: a review from the swarm engineering perspective.
\newblock \emph{Swarm Intelligence}, 7(1), 1--41.

\bibitem[{Burgard et~al.(2000)Burgard, Moors, Fox, Simmons, and
  Thrun}]{Ref:Burgard00}
Burgard, W., Moors, M., Fox, D., Simmons, R., and Thrun, S. (2000).
\newblock Collaborative multi-robot exploration.
\newblock In \emph{{IEEE} International Conference on Robotics and Automation},
  volume~1.

\bibitem[{B\"{u}rger et~al.(2012)B\"{u}rger, Notarstefano, Bullo, and
  Allg\"{o}wer}]{Ref:Bullo12}
B\"{u}rger, M., Notarstefano, G., Bullo, F., and Allg\"{o}wer, F. (2012).
\newblock A distributed simplex algorithm for degenerate linear programs and
  multi-agent assignments.
\newblock \emph{Automatica}, 48, 2298--2304.

\bibitem[{Cao et~al.(2013)Cao, Yu, Ren, and Chen}]{Ref:Cao13}
Cao, Y., Yu, W., Ren, W., and Chen, G. (2013).
\newblock An overview of recent progress in the study of distributed
  multi-agent coordination.
\newblock \emph{IEEE Transactions on Industrial Informatics}, 9(1), 427--438.

\bibitem[{Chalkiadakis and Boutilier(2003)}]{Ref:Chalkiadakis03}
Chalkiadakis, G. and Boutilier, C. (2003).
\newblock Coordination in multiagent reinforcement learning: A bayesian
  approach.
\newblock In \emph{Proceedings of the second international joint conference on
  Autonomous agents and multiagent systems}.

\bibitem[{Chung et~al.(2013)Chung, Bandyopadhyay, Chang, and
  Hadaegh}]{Ref:Chung12}
Chung, S.J., Bandyopadhyay, S., Chang, I., and Hadaegh, F.Y. (2013).
\newblock Phase synchronization control of complex networks of {L}agrangian
  systems on adaptive digraphs.
\newblock \emph{Automatica}, 49(5), 1148--1161.

\bibitem[{Cort\'{e}s et~al.(2006)Cort\'{e}s, Martinez, and
  Bullo}]{Ref:Cortes06}
Cort\'{e}s, J., Martinez, S., and Bullo, F. (2006).
\newblock \emph{{IEEE} Trans. Autom. Control}, 51(8), 1289--1298.

\bibitem[{Cort\'{e}s et~al.(2004)Cort\'{e}s, Martinez, Karatas, and
  Bullo}]{Ref:Bullo04}
Cort\'{e}s, J., Martinez, S., Karatas, T., and Bullo, F. (2004).
\newblock Coverage control for mobile sensing networks.
\newblock \emph{{IEEE} Trans. Robotics and Automation}, 20(2).

\bibitem[{Desaraju and How(2012)}]{Ref:How12}
Desaraju, V.R. and How, J.P. (2012).
\newblock Decentralized path planning for multi-agent teams with complex
  constraints.
\newblock \emph{Autonomous Robots}, 32(4), 385--403.

\bibitem[{Dias(2004)}]{Ref:Dias04}
Dias, M.B. (2004).
\newblock Traderbots: A new paradigm for robust and efficient multirobot
  coordination in dynamic environments.
\newblock \emph{Robotics Institute}, 153.

\bibitem[{Dunbar and Murray(2002)}]{Ref:Dunbar02}
Dunbar, W.B. and Murray, R.M. (2002).
\newblock Model predictive control of coordinated multi-vehicle formations.
\newblock In \emph{{IEEE Conf. Decision Control}}, volume~4.

\bibitem[{Earl and D'Andrea(2002)}]{Ref:Earl02modeling}
Earl, M.G. and D'Andrea, R. (2002).
\newblock Modeling and control of a multi-agent system using mixed integer
  linear programming.
\newblock In \emph{{IEEE} Conf. Decision Control}, volume~1.

\bibitem[{Eich et~al.(2014)Eich, Hartanto, Kasperski, Natarajan, and
  Wollenberg}]{Ref:Eich14}
Eich, M., Hartanto, R., Kasperski, S., Natarajan, S., and Wollenberg, J.
  (2014).
\newblock Towards coordinated multirobot missions for lunar sample collection
  in an unknown environment.
\newblock \emph{Journal of Field Robotics}, 31(1).

\bibitem[{Feddema et~al.(2002)Feddema, Lewis, and Schoenwald}]{Ref:Feddema02}
Feddema, J.T., Lewis, C., and Schoenwald, D.A. (2002).
\newblock Decentralized control of cooperative robotic vehicles: theory and
  application.
\newblock \emph{{IEEE} Trans. Robotics and Automation}, 18(5), 852--864.

\bibitem[{Fredslund and Mataric(2002)}]{Ref:Fredslund02}
Fredslund, J. and Mataric, M.J. (2002).
\newblock A general algorithm for robot formations using local sensing and
  minimal communication.
\newblock \emph{{IEEE} Trans. Robotics and Automation}, 18(5), 837--846.

\bibitem[{Garin and Schenato(2010)}]{Ref:Garin10}
Garin, F. and Schenato, L. (2010).
\newblock A survey on distributed estimation and control applications using
  linear consensus algorithms.
\newblock In \emph{Networked Control Systems}, 75--107. Springer.

\bibitem[{Gerkey and Mataric(2002)}]{Ref:Gerkey02}
Gerkey, B.P. and Mataric, M.J. (2002).
\newblock Sold!: auction methods for multirobot coordination.
\newblock \emph{{IEEE} Trans. Robotics and Automation}, 18(5), 758--768.

\bibitem[{Gittins(1979)}]{Ref:Gittins79}
Gittins, J.C. (1979).
\newblock Bandit processes and dynamic allocation indices.
\newblock \emph{Journal of the Royal Statistical Society. Series B
  (Methodological)}, 148--177.

\bibitem[{Hereford(2011)}]{Ref:Hereford11}
Hereford, J.M. (2011).
\newblock Analysis of {BEECLUST} swarm algorithm.
\newblock In \emph{{IEEE} Symp. Swarm Intell.}

\bibitem[{Khatib(1986)}]{Ref:Khatib86}
Khatib, O. (1986).
\newblock Real-time obstacle avoidance for manipulators and mobile robots.
\newblock \emph{Int. J. Robotics Research}, 5(1), 90--98.

\bibitem[{King et~al.(2003)King, Pretty, and Gosine}]{Ref:King03}
King, J., Pretty, R.K., and Gosine, R.G. (2003).
\newblock Coordinated execution of tasks in a multiagent environment.
\newblock \emph{IEEE Transactions on Systems, Man, and Cybernetics - Part A:
  Systems and Humans}, 33(5), 615--619.

\bibitem[{Koditschek and Rimon(1990)}]{Ref:Koditschek90}
Koditschek, D.E. and Rimon, E. (1990).
\newblock Robot navigation functions on manifolds with boundary.
\newblock \emph{Advances in applied mathematics}, 11(4), 412--442.

\bibitem[{Kress-Gazit et~al.(2008)Kress-Gazit, Conner, Choset, Rizzi, and
  Pappas}]{Ref:KressGazit08}
Kress-Gazit, H., Conner, D.C., Choset, H., Rizzi, A.A., and Pappas, G.J.
  (2008).
\newblock Courteous cars.
\newblock \emph{IEEE Robotics Automation Magazine}, 15(1), 30--38.

\bibitem[{Landgren et~al.(2016)Landgren, Srivastava, and
  Leonard}]{Ref:Leonard16}
Landgren, P., Srivastava, V., and Leonard, N.E. (2016).
\newblock On distributed cooperative decision-making in multiarmed bandits.
\newblock In \emph{{IEEE European Control Conf.}}

\bibitem[{Le~Ny et~al.(2008)Le~Ny, Dahleh, and Feron}]{Ref:Le08}
Le~Ny, J., Dahleh, M., and Feron, E. (2008).
\newblock Multi-{UAV} dynamic routing with partial observations using restless
  bandit allocation indices.
\newblock In \emph{{IEEE American Control Conf.}}

\bibitem[{Leonard et~al.(2010)Leonard, Paley, Davis, Fratantoni, Lekien, and
  Zhang}]{Ref:Leonard10b}
Leonard, N.E., Paley, D.A., Davis, R.E., Fratantoni, D.M., Lekien, F., and
  Zhang, F. (2010).
\newblock Coordinated control of an underwater glider fleet in an adaptive
  ocean sampling field experiment in monterey bay.
\newblock \emph{Journal of Field Robotics}, 27(6), 718--740.

\bibitem[{Li and Rus(2006)}]{Ref:Li2006global}
Li, Q. and Rus, D. (2006).
\newblock Global clock synchronization in sensor networks.
\newblock \emph{IEEE Transactions on computers}, 55(2), 214--226.

\bibitem[{Liu and Goldsmith(2004)}]{Ref:Liu04}
Liu, X. and Goldsmith, A. (2004).
\newblock Kalman filtering with partial observation losses.
\newblock In \emph{{IEEE} Conf. Decision Control}, volume~4.

\bibitem[{Liu and Nejat(2016)}]{Ref:Liu16}
Liu, Y. and Nejat, G. (2016).
\newblock Multirobot cooperative learning for semiautonomous control in urban
  search and rescue applications.
\newblock \emph{Journal of Field Robotics}, 33(4), 512--536.

\bibitem[{Lumelsky and Harinarayan(1997)}]{Ref:Lumelsky97}
Lumelsky, V. and Harinarayan, K. (1997).
\newblock Decentralized motion planning for multiple mobile robots: The
  cocktail party model.
\newblock \emph{Autonomous Robots}, 4(1), 121--135.

\bibitem[{Lynch(1997)}]{Ref:Lynch97}
Lynch, N.A. (1997).
\newblock \emph{Distributed Algorithms}.
\newblock Morgan Kaufmann.

\bibitem[{Marshall et~al.(2004)Marshall, Broucke, and Francis}]{Ref:Marshall04}
Marshall, J.A., Broucke, M.E., and Francis, B.A. (2004).
\newblock Formations of vehicles in cyclic pursuit.
\newblock \emph{{IEEE} Trans. Autom. Control}, 49(11), 1963--1974.

\bibitem[{Mathews et~al.(2017)Mathews, Christensen, O'Grady, Mondada, and
  Dorigo}]{Ref:Dorigo17}
Mathews, N., Christensen, A.L., O'Grady, R., Mondada, F., and Dorigo, M.
  (2017).
\newblock Mergeable nervous systems for robots.
\newblock \emph{Nature Communications}, 8(1), 439.

\bibitem[{Menon et~al.(2004)Menon, Sweriduk, and Bilimoria}]{Ref:Menon04}
Menon, P.K., Sweriduk, G.D., and Bilimoria, K.D. (2004).
\newblock New approach for modeling, analysis, and control of air traffic flow.
\newblock \emph{J. Guid. Control Dyn.}, 27(5).

\bibitem[{Mesbahi and Hadaegh(1999)}]{Ref:Mesbahi99formation}
Mesbahi, M. and Hadaegh, F.Y. (1999).
\newblock Formation flying control of multiple spacecraft via graphs, matrix
  inequalities, and switching.
\newblock In \emph{{IEEE Int. Conf.\ on Control Applications}}, volume~2.

\bibitem[{Mesquita et~al.(2008)Mesquita, Hespanha, and
  \r{A}str\"{o}m}]{Ref:Hespanha08}
Mesquita, A.R., Hespanha, J.P., and \r{A}str\"{o}m, K. (2008).
\newblock Optimotaxis: A stochastic multi-agent optimization procedure with
  point measurements.
\newblock In \emph{Proc. Hybrid Systems: Computation and Control}.

\bibitem[{Morgan et~al.(2016)Morgan, Subramanian, Chung, and
  Hadaegh}]{Ref:Morgan15_SATO}
Morgan, D., Subramanian, G.P., Chung, S.J., and Hadaegh, F.Y. (2016).
\newblock Swarm assignment and trajectory optimization using variable-swarm,
  distributed auction assignment and sequential convex programming.
\newblock \emph{Int. J. Robotics Research}, 35, 1261--1285.

\bibitem[{Ogren et~al.(2002)Ogren, Egerstedt, and Hu}]{Ref:Ogren02}
Ogren, P., Egerstedt, M., and Hu, X. (2002).
\newblock A control lyapunov function approach to multiagent coordination.
\newblock \emph{{IEEE} Trans. Robotics and Automation}, 18(5), 847--851.

\bibitem[{Oh et~al.(2015)Oh, Park, and Ahn}]{Ref:Oh15survey}
Oh, K.K., Park, M.C., and Ahn, H.S. (2015).
\newblock A survey of multi-agent formation control.
\newblock \emph{Automatica}, 53, 424--440.

\bibitem[{Olfati-Saber(2006)}]{Ref:Saber06}
Olfati-Saber, R. (2006).
\newblock Flocking for multi-agent dynamic systems: algorithms and theory.
\newblock \emph{{IEEE} Trans. Autom. Control}, 51(3), 401--420.

\bibitem[{Olson et~al.(2012)Olson, Strom, Morton, Richardson, Ranganathan,
  Goeddel, Bulic, Crossman, and Marinier}]{Ref:Olson12}
Olson, E., Strom, J., Morton, R., Richardson, A., Ranganathan, P., Goeddel, R.,
  Bulic, M., Crossman, J., and Marinier, B. (2012).
\newblock Progress toward multi-robot reconnaissance and the {MAGIC} 2010
  competition.
\newblock \emph{Journal of Field Robotics}, 29(5), 762--792.

\bibitem[{Omidshafiei et~al.(2015)Omidshafiei, Agha-Mohammadi, Amato, and
  How}]{Ref:Ali15}
Omidshafiei, S., Agha-Mohammadi, A.A., Amato, C., and How, J.P. (2015).
\newblock Decentralized control of partially observable markov decision
  processes using belief space macro-actions.
\newblock In \emph{{IEEE} International Conference on Robotics and Automation}.

\bibitem[{Pavone et~al.(2011)Pavone, Arsie, Frazzoli, and Bullo}]{Ref:Pavone11}
Pavone, M., Arsie, A., Frazzoli, E., and Bullo, F. (2011).
\newblock Distributed algorithms for environment partitioning in mobile robotic
  networks.
\newblock \emph{{IEEE} Trans. Autom. Control}, 56(8), 1834--1848.

\bibitem[{Pavone et~al.(2012)Pavone, Smith, Frazzoli, and Rus}]{Ref:Pavone11b}
Pavone, M., Smith, S.L., Frazzoli, E., and Rus, D. (2012).
\newblock Robotic load balancing for mobility-on-demand systems.
\newblock \emph{Int. J. Robotics Research}, 31(7), 839--854.

\bibitem[{Rabbat and Nowak(2004)}]{Ref:Rabbat04}
Rabbat, M. and Nowak, R. (2004).
\newblock Distributed optimization in sensor networks.
\newblock In \emph{Proceedings of the 3rd international symposium on
  Information processing in sensor networks}.

\bibitem[{Ren et~al.(2007)Ren, Beard, and Atkins}]{Ref:Ren07}
Ren, W., Beard, R.W., and Atkins, E.M. (2007).
\newblock Information consensus in multivehicle cooperative control.
\newblock \emph{{IEEE} Control Syst. Mag.}, 27(2), 71--82.

\bibitem[{Reveliotis and Roszkowska(2011)}]{Ref:Reveliotis11}
Reveliotis, S.A. and Roszkowska, E. (2011).
\newblock Conflict resolution in free-ranging multivehicle systems: A resource
  allocation paradigm.
\newblock \emph{{IEEE} Trans. Robotics}, 27(2), 283--296.

\bibitem[{Richards et~al.(2002)Richards, Schouwenaars, How, and
  Feron}]{Ref:How02}
Richards, A., Schouwenaars, T., How, J., and Feron, E. (2002).
\newblock Spacecraft trajectory planning with avoidance constraints using
  mixed-integer linear programming.
\newblock \emph{J. Guid. Control Dyn.}, 25(4).

\bibitem[{Richards and How(2007)}]{Ref:How07MPC}
Richards, A. and How, J.P. (2007).
\newblock Robust distributed model predictive control.
\newblock \emph{International Journal of Control}, 80(9), 1517--1531.

\bibitem[{Rossi and Pavone(2014)}]{Ref:Rossi14}
Rossi, F. and Pavone, M. (2014).
\newblock On the fundamental limitations of performance for distributed
  decision-making in robotic networks.
\newblock In \emph{Proc. IEEE Conf. on Decision and Control}.

\bibitem[{Rossi et~al.(2018)Rossi, Zhang, Hindy, and Pavone}]{Ref:Rossi17a}
Rossi, F., Zhang, R., Hindy, Y., and Pavone, M. (2018).
\newblock Routing autonomous vehicles in congested transportation networks:
  structural properties and coordination algorithms.
\newblock \emph{Autonomous Robots}.
\newblock In Press.

\bibitem[{Rubenstein et~al.(2014)Rubenstein, Cornejo, and
  Nagpal}]{Ref:Nagpal14}
Rubenstein, M., Cornejo, A., and Nagpal, R. (2014).
\newblock Programmable self-assembly in a thousand-robot swarm.
\newblock \emph{Science}, 345(6198), 795--799.

\bibitem[{Rus et~al.(1995)Rus, Donald, and Jennings}]{Ref:Rus1995moving}
Rus, D., Donald, B., and Jennings, J. (1995).
\newblock Moving furniture with teams of autonomous robots.
\newblock In \emph{IEEE/RSJ Int.\ Conf.\ on Intelligent Robots \& Systems},
  volume~1.

\bibitem[{Sahai and Mitter(2006)}]{Ref:Sahai06}
Sahai, A. and Mitter, S. (2006).
\newblock The necessity and sufficiency of anytime capacity for stabilization
  of a linear system over a noisy communication link - part i: Scalar systems.
\newblock \emph{IEEE Trans. on Information Theory}, 52(8), 3369--3395.

\bibitem[{Scattolini(2009)}]{Ref:Scattolini09}
Scattolini, R. (2009).
\newblock Architectures for distributed and hierarchical model predictive
  control - a review.
\newblock \emph{Journal of Process Control}, 19(5), 723 -- 731.

\bibitem[{Schouwenaars et~al.(2006)Schouwenaars, Stubbs, Paduano, and
  Feron}]{Ref:How06}
Schouwenaars, T., Stubbs, A., Paduano, J., and Feron, E. (2006).
\newblock Multivehicle path planning for nonline-of-sight communication.
\newblock \emph{Journal of Field Robotics}, 23(3-4), 269--290.

\bibitem[{Schwager et~al.(2009)Schwager, McLurkin, Slotine, and
  Rus}]{Ref:Schwager09}
Schwager, M., McLurkin, J., Slotine, J.J., and Rus, D. (2009).
\newblock From theory to practice: Distributed coverage control experiments
  with groups of robots.
\newblock In \emph{Experimental Robotics}.

\bibitem[{Sepulchre et~al.(2007)Sepulchre, Paley, and Leonard}]{Ref:Leonard07}
Sepulchre, R., Paley, D.A., and Leonard, N.E. (2007).
\newblock Stabilization of planar collective motion: All-to-all communication.
\newblock \emph{{IEEE} Trans. Autom. Control}, 52(5), 811--824.

\bibitem[{Sharon et~al.(2015)Sharon, Stern, Felner, and
  Sturtevant}]{Ref:Sturtevant15}
Sharon, G., Stern, R., Felner, A., and Sturtevant, N.R. (2015).
\newblock Conflict-based search for optimal multi-agent pathfinding.
\newblock \emph{Artificial Intelligence}, 219(Supplement C), 40 -- 66.

\bibitem[{Shehory and Kraus(1998)}]{Ref:Shehory98}
Shehory, O. and Kraus, S. (1998).
\newblock Methods for task allocation via agent coalition formation.
\newblock \emph{Artificial Intelligence}, 101(1), 165 -- 200.

\bibitem[{Str{\"o}mbom et~al.(2014)Str{\"o}mbom, Mann, Wilson, Hailes, Morton,
  Sumpter, and King}]{Ref:Strobom14}
Str{\"o}mbom, D., Mann, R.P., Wilson, A.M., Hailes, S., Morton, A.J., Sumpter,
  D.J., and King, A.J. (2014).
\newblock Solving the shepherding problem: heuristics for herding autonomous,
  interacting agents.
\newblock \emph{Journal of the Royal Society Interface}, 11(100), 20140719.

\bibitem[{Tanner et~al.(2007)Tanner, Jadbabaie, and Pappas}]{Ref:Tanner07}
Tanner, H.G., Jadbabaie, A., and Pappas, G.J. (2007).
\newblock Flocking in fixed and switching networks.
\newblock \emph{{IEEE} Trans. Autom. Control}, 52(5), 863--868.

\bibitem[{Tsitsiklis et~al.(1986)Tsitsiklis, Bertsekas, and
  Athans}]{Ref:Tsitsiklis86}
Tsitsiklis, J.N., Bertsekas, D.P., and Athans, M. (1986).
\newblock Distributed asynchronous deterministic and stochastic gradient
  optimization algorithms.
\newblock \emph{{IEEE} Trans. Autom. Control}, 31(9), 803 -- 812.

\bibitem[{Turpin et~al.(2014)Turpin, Michael, and Kumar}]{Ref:Kumar14}
Turpin, M., Michael, N., and Kumar, V. (2014).
\newblock {CAPT}: Concurrent assignment and planning of trajectories for
  multiple robots.
\newblock \emph{Int. J. Robotics Research}, 33(1), 98--112.

\bibitem[{van~den Berg et~al.(2008)van~den Berg, Lin, and
  Manocha}]{Ref:vabdenBerg08}
van~den Berg, J., Lin, M., and Manocha, D. (2008).
\newblock Reciprocal velocity obstacles for real-time multi-agent navigation.
\newblock In \emph{{IEEE} International Conference on Robotics and Automation}.

\bibitem[{Von~Stryk and Bulirsch(1992)}]{Ref:VonStryk92}
Von~Stryk, O. and Bulirsch, R. (1992).
\newblock Direct and indirect methods for trajectory optimization.
\newblock \emph{Annals of operations research}, 37(1), 357--373.

\bibitem[{Weigel et~al.(2002)Weigel, Gutmann, Dietl, Kleiner, and
  Nebel}]{Ref:Weigel02}
Weigel, T., Gutmann, J.S., Dietl, M., Kleiner, A., and Nebel, B. (2002).
\newblock {CS Freiburg}: coordinating robots for successful soccer playing.
\newblock \emph{{IEEE} Trans. Robotics and Automation}, 18(5), 685--699.

\bibitem[{Werfel et~al.(2014)Werfel, Petersen, and
  Nagpal}]{Ref:Nagpal14termite}
Werfel, J., Petersen, K., and Nagpal, R. (2014).
\newblock Designing collective behavior in a termite-inspired robot
  construction team.
\newblock \emph{Science}, 343(6172), 754--758.

\bibitem[{Wu and Zhang(2012)}]{Ref:Wu12}
Wu, W. and Zhang, F. (2012).
\newblock Robust cooperative exploration with a switching strategy.
\newblock \emph{{IEEE} Trans. Robotics}, 28(4), 828--839.

\bibitem[{Xu et~al.(2013)Xu, Fitch, Underwood, and Sukkarieh}]{Ref:Zhe13}
Xu, Z., Fitch, R., Underwood, J., and Sukkarieh, S. (2013).
\newblock Decentralized coordinated tracking with mixed discrete-continuous
  decisions.
\newblock \emph{Journal of Field Robotics}, 30(5), 717--740.

\bibitem[{Yu et~al.(2008)Yu, LaValle, and Liberzon}]{Ref:Yu08rendezvous}
Yu, J., LaValle, S.M., and Liberzon, D. (2008).
\newblock Rendezvous without coordinates.
\newblock In \emph{{IEEE} Conf. Decision Control}.

\bibitem[{Yu et~al.(2014)Yu, Schwager, and Rus}]{Ref:Rus14}
Yu, J., Schwager, M., and Rus, D. (2014).
\newblock Correlated orienteering problem and its application to informative
  path planning for persistent monitoring tasks.
\newblock In \emph{IEEE/RSJ Int.\ Conf.\ on Intelligent Robots \& Systems}.

\bibitem[{Zavlanos et~al.(2007)Zavlanos, Jadbabaie, and
  Pappas}]{Ref:Zavlanos2007flocking}
Zavlanos, M.M., Jadbabaie, A., and Pappas, G.J. (2007).
\newblock Flocking while preserving network connectivity.
\newblock In \emph{{IEEE} Conf. Decision Control}.

\bibitem[{Zhao et~al.(2011)Zhao, Ramakrishnan, and Kumar}]{Ref:MKumar11}
Zhao, S., Ramakrishnan, S., and Kumar, M. (2011).
\newblock Density-based control of multiple robots.
\newblock In \emph{{IEEE American Control Conf.}}

\bibitem[{Zhou et~al.(2017)Zhou, Wang, Bandyopadhyay, and
  Schwager}]{Ref:Zhou17}
Zhou, D., Wang, Z., Bandyopadhyay, S., and Schwager, M. (2017).
\newblock Fast, on-line collision avoidance for dynamic vehicles using buffered
  voronoi cells.
\newblock \emph{IEEE Robotics and Automation Letters}, 2(2), 1047--1054.

\end{thebibliography}
}

\end{document}